\documentclass[letterpaper, 10 pt, conference]{ieeeconf}  

\IEEEoverridecommandlockouts                              

\usepackage{graphicx} 
\usepackage{tikz}
\usepackage{amsmath} 
\usepackage{comment}
\usepackage{float}
\usepackage{color, soul}
\usepackage{tabularx}
\usepackage{xcolor}
\usepackage{multirow}
\usepackage{booktabs}
\usepackage{makecell}

\title{\LARGE \bf Digital twin and extended reality for teleoperation of the electric vehicle battery disassembly}

\author{Tero Kaarlela$^{1,4}$ and Sami Salo$^{2}$ and Jose Outeiro$^{3}$
\thanks{$^{1}$Knight Foundation School of Computing and Information Sciences, Florida International University, Miami, Florida, USA.
        {\tt\small tkaarlel@fiu.edu}}%
\thanks{$^{2}$Centria University of Applied Sciences, Department of Production Technology, Vierimaantie 7, Ylivieska, 84100, Finland
        {\tt\small sami.salo@centria.fi}}%
\thanks{$^{3}$Digital Engineering for Advanced Manufacturing Laboratory (DEAM Lab), Center for Precision Metrology, Department of Mechanical Engineering and Engineering Science, University of North Carolina at Charlotte, 9201 University City Blvd., Charlotte 28223, NC, USA
       {\tt\small jc.outeiro@charlotte.edu}}%
\thanks{$^{4}$Materials and Mechanical Engineering, Faculty of Technology, University of Oulu, Oulu, Finland.
        {\tt\small tero.kaarlela@oulu.fi}}%
}

\begin{document}

\maketitle
\thispagestyle{empty}
\pagestyle{empty}

\begin{abstract}
Disassembling and sorting Electric Vehicle Batteries (EVBs) supports a sustainable transition to electric vehicles by enabling a closed-loop supply chain. Currently, the manual disassembly process exposes workers to hazards, including electrocution and toxic chemicals. We propose a teleoperated system for the safe disassembly and sorting of EVBs. A human-in-the-loop can create and save disassembly sequences for unknown EVB types, enabling future automation. An RGB camera aligns the physical and digital twins of the EVB, and the digital twin of the robot is based on the Robot Operating System (ROS) middleware. This hybrid approach combines teleoperation and automation to improve safety, adaptability, and efficiency in EVB disassembly and sorting. The economic contribution is realized by reducing labor dependency and increasing throughput in battery recycling. An online pilot study was set up to evaluate the usability of the presented approach, and the results demonstrate the potential as a user-friendly solution.
\end{abstract}

\section{INTRODUCTION}
\label{sec:introduction}

As the adoption of electric vehicles is increasing, efficient processes for effective material recovery from end-of-life EVBs are required to minimize the carbon footprint of EVB manufacturing and mitigate the environmental and social impacts of mining~\cite{fernandeznavarro2012,ramos2008}. Complex assemblies consisting of diverse components require effective methods for disassembly and sorting~\cite{Priyono2016}. 

\subsection{Motivation}
\label{subsec:motivation}
Manual EVB disassembly exposes workers to multiple hazards, including chemical leaks, thermal runaways, and high-voltage electric shocks, which can occur due to accidental short circuits during recycling, repurposing, or reuse. To mitigate these risks, professional disassembly typically requires at least two workers to ensure safety and provide immediate assistance in case of an incident. However, this approach remains labor-intensive and inherently hazardous, necessitating safer and more efficient alternatives.

Although an autonomous dismantling and sorting system avoids exposing human workers to the mentioned risks, developing an autonomous system for dismantling and sorting EVBs is challenging. The diversity in the physical qualities of different battery types and the lack of design for disassembly require cognitive capabilities and dexterity comparable to those of humans. These requirements set a high demand for flexibility and reconfigurability for autonomous disassembly~\cite{Kaarlela2024}, and therefore, EVB disassembly is typically manual work~\cite{Zang2022}.

Researchers have proposed the introduction of an industrial robot arm as the disassembly mechanism to enable flexibility and reconfigurability~\cite{Villagrossi2023}. An industrial robot provides a high-level controller for flexible reprogramming to a diverse selection of EVBs. The automatic tool changer enables a diverse selection of tooling for the detachment and handling tasks required during disassembly. The detaching and sorting of the components, such as electrical connectors and screws, is challenging, requiring identifying, locating, and mechanical action. In addition, trajectory and strategy planning require cognitive capabilities to decide the optimal approach direction before mechanical action, such as gripping or unscrewing, and to decide the optimal disassembly sequence.

\subsection{State-of-the-art about teleoperation}
\label{subsec:stateoftheart}
Teleoperation has been applied to enable safe working in hazardous environments through physical barriers by using remote-controlled manipulators. Goertz~\cite{Goertz1952} applied teleoperation to sample radioactive materials at a nuclear plant in the 1950s. Although robot teleoperation in radioactive environments remains an active research topic~\cite{Szczurek2023}, teleoperation has been widely applied, for example, in construction~\cite{Ito2019} and education~\cite{poysari2023}.

Hathaway et al. presented a teleoperated approach to EVB disassembly~\cite{Hathaway2023}. They used a dedicated slave device to control the master device during the disassembly tasks. Their study noted that using identical slave and master devices is more convenient for the teleoperator than using a slave device with a deviating configuration. In their work, two camera views were provided to the teleoperator to perform handling, cutting, and unscrewing tasks.  

Digital Twins (DTs) have been applied for teleoperation to enable skill training in programming and controlling industrial and collaborative robots. Garg et al.~\cite{Garg2021} presented a framework to create and validate trajectories for Fanuc robots. Xin et al.~\cite{Xin2021} proposed a multisource DT that enables real-time sensing of the surrounding environment using an RGB-D camera and a force sensor.

Alternative approaches, such as Human-Robot Collaboration (HRC), have been proposed for EVB disassembly~\cite{Villagrossi2023}. HRC enables the use of a mixture of human workers and collaborative robots, combining the strengths of humans and robots. However, utilizing HRC exposes the human worker to the hazards of the disassembly environment.

The need for six-degrees-of-freedom (6DoF) pose estimation in Extended Reality (XR) and robotic applications has increased recently. Pose estimation requires removing the target from other objects and defining the target orientation and position. MegaPose~\cite{labbe2022} is used in the present research work to update the 6DoF pose of the physical EVB to the DT. Instead of training a dataset for each target object, MegaPose enables using a CAD model of the target object and an RGB or RGB-D camera for pose estimation.

\subsection{Scientific contribution}
\label{subsec:scientificcontribution}
This publication presents a concept of DT-based teleoperation for disassembly and sorting EVBs. Compared to Hathaway et al. and Xin et al., our proposed approach introduces a hybrid methodology combining teleoperation with digital twins for trajectory planning and collision avoidance. Unlike previous works that rely on identical master-slave device configurations or lack real-time updates, our system leverages an XR-based user interface and a ROS-integrated DT for enhanced adaptability and safety. Table~\ref{tab:comparison} summarizes the features of the previous solutions and the proposed approach.

\begin{table}[!tb]
\centering
\caption{Comparison of related works in EVB disassembly.}
\resizebox{\columnwidth}{!}{%
\begin{tabular}{|l||c||c||c||c|}
\hline
\textbf{Approach} & \textbf{Teleoperation} & \textbf{DT} & \textbf{Automation} & \textbf{Collision Avoidance} \\\hline
Hathaway et al.~\cite{Hathaway2023} & Yes & No & No & No \\\hline
Xin et al.~\cite{Xin2021} & Yes & Yes & No & No \\\hline
\textbf{Proposed Approach} & Yes & Yes & Semi-automated & Yes \\\hline
\end{tabular}%
}
\label{tab:comparison}
\end{table}

The presented concept does not provide a fully autonomous EVB disassembly solution for the entire range of battery types. Instead, it introduces a hybrid approach that combines teleoperation and automation. A teleoperator programs the disassembly sequence for a specific EVB type once, leveraging the flexibility of the robot's high-level controller and tool-changing capabilities. Once this programming is complete, the system can automatically disassemble and sort similar EVBs without additional programming. 

This approach combines the strengths of automated systems and human cognitive capabilities while ensuring safe and efficient disassembly for a diverse range of EVBs. Integrating human expertise in the initial programming phase allows the system to handle the challenges posed by the diversity and complexity of EVB designs. Subsequent automated operation enhances efficiency and scalability. Safety is improved as the human worker monitors and controls the hazardous disassembly process from a safe location.

In the presented approach, the DT guides the teleoperator by providing visual aids to generate collision-free trajectories. The Robot Sensor Interface (RSI) software add-on installed in the robot controller enables bidirectional communication required for teleoperation; the communication layer is based on TCP/IP. 

\section{Proposed concept for EVB disassembly}
The proposed concept for EVB disassembly integrates physical and digital twins to enable an efficient and safe EVB disassembly process. The physical twin consisting of an industrial robot and an EVB setup is presented in Section~\ref{subsec:physicaltwin}, the DT-based ROS kinematic model and XR user interface are presented in Section~\ref{subsec:digitaltwin}, and the details of the communication layer are given in Section~\ref{subsec:bidirectionalcommunication}.

\subsection{Physical twin}
\label{subsec:physicaltwin}
The physical twins are located in Centria's laboratory for robotized EVB disassembly and sorting. The physical twins are schematically represented in Figure 4 and comprise an industrial robot (Kuka KR10 R1100) installed to hang downward on a linear track from a steel frame bolted to the concrete floor and an EVB for disassembly positioned below the robot installation. 

The robot is installed hanging down from a floor-mounted steel frame to maximize workspace and to provide unobstructed access to the EVB. This setup minimizes floor obstruction and ensures flexibility in tool positioning. The steel frame can be easily relocated into a high-cube sea container, enabling rapid deployment in an industrially relevant environment while providing an isolated environment for hazardous disassembly tasks. 

The mentioned setup enables seven degrees of freedom and a working area of 4 x 2 meters. The robot is equipped with an ATC, consisting of a master adapter installed on the robot flange and a slave adapter installed on each tool. A stationary tool holder for storing the resting tools is mounted in one corner of the steel frame structure. Tools enable common EVB disassembly and sorting tasks such as unscrewing, handling, and detaching wiring assemblies and connectors.  

\begin{figure}[!tb]
  \centering
  \framebox{\parbox{3in}{\includegraphics[width=3in]{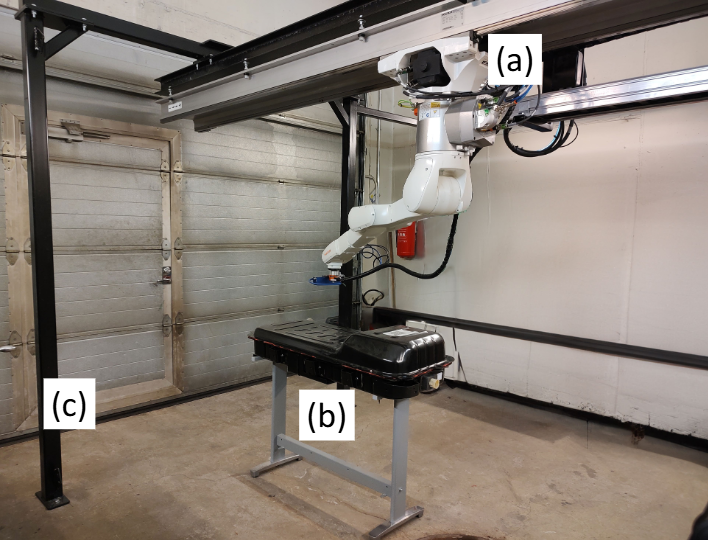}}}
  \caption{Physical twin consisting of (a) robot and linear track, (b) EVB, (c) steel frame.}
  \label{fig:phystwin}
\end{figure}

The physical twin of the EVB is a Plug-in Hybrid Electric Vehicle (PHEV) battery. The EVB consists of thermally managed pouch cells controlled by a Battery Monitoring System (BMS). The nominal capacity of the EVB is 12.5 kWh, and the nominal voltage is 400 volts. External dimensions are 1.5 x 0.8 x 0.5 meters, and weight is 174 kgs. The physical twin is illustrated in Figure~\ref{fig:phystwin} 

\subsection{Digital twin}
\label{subsec:digitaltwin}

The robot's DT was created using the Robot Operating System (ROS). The ROS middleware connects the XR user interface and the physical robot, enabling accurate simulation and planning of robot trajectories and providing the required forward and inverse kinematics.  
The XR user interface is built using the Unity game engine to enable a realistic environment for the teleoperator. Utilizing the DT, the user interface guides the teleoperator in programming collision-free trajectories. The physical disassembly and sorting environment (the physical twin) was digitized using an industrial 3D scanner, and the resulting point cloud served as a template to extrude stationary objects, such as walls, floors, and pillars. The robot station frame was then positioned in the digital model using Blender software. The robot driver and kinematics are based on the kuka\_robot\_descriptions ROS package~\cite{kroshukukadescription}.

To create a kinematic model for the robot, the bone structure in Blender was used to tie the robot joints together and define the angular constraints for each joint. The created digital version of the disassembly environment and the required tools were imported into Unity to add functionality and bidirectional communication between the DT and the corresponding physical twin.

\begin{figure}[!tb]
  \centering
  \framebox{\parbox{3in}{\includegraphics[width=3in]{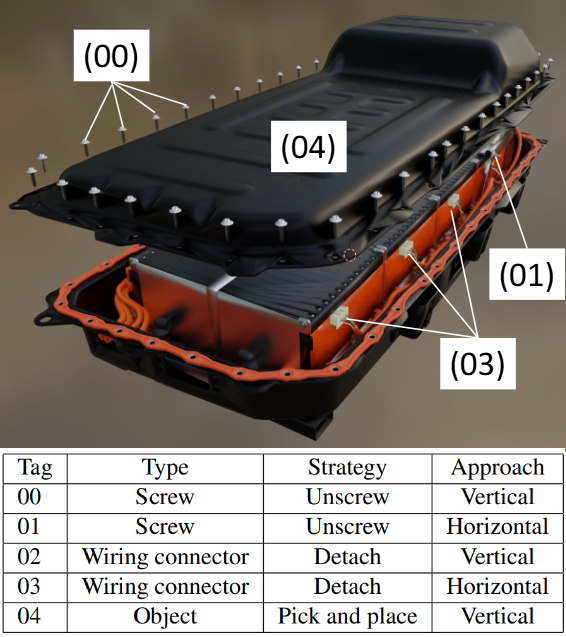}}}
  \caption{Illustration of the tags included in the CAD model.}
  \label{fig:batterycad}
\end{figure}

\begin{figure}[!tb]
  \centering
  \framebox{\parbox{3in}{\includegraphics[width=3in]{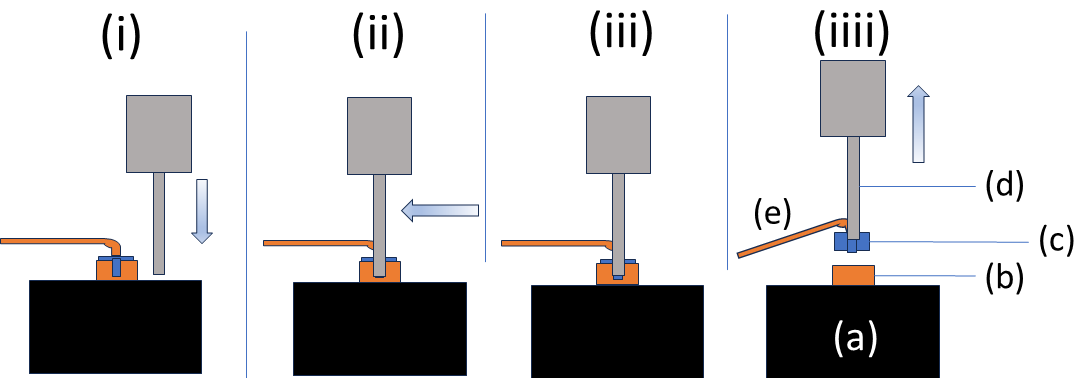}}}
  \caption{Connector detaching strategy (a) BMS module, (b) connector socket, (c) connector plug, (d) gripper fingers. Phases of detaching: (i) vertical approach from the side opposite to the wiring harness exit direction, (ii) approach over the latch mechanism, (iii) close the gripper to unlatch the connector, (iiii) pull the plug out of the socket.}
  \label{fig:strategy}
\end{figure}

The CAD model of the EVB includes all the components of the assembly, such as the upper and lower halves of the enclosure, the enclosure seal, the battery modules, the wiring harness, the fuse and contactor unit, the BMS, radiator for thermal management and all the screws required to mount the components to the EVB assembly. Each component includes a tag that defines the strategy, the orientation of the approach, and the tool necessary for detaching. For example, if the teleoperator clicks on a horizontally mounted screw component of the EVB, unscrew tool is activated and positioned at the approach point. After initial positioning, a subprogram is run to trigger the unscrew cycle. The robot's linear speed during unscrewing is synchronized to the screw rotation speed using tools internal speed signal. Tags and corresponding strategies are detailed in Figure~\ref{fig:batterycad}. The strategy for detaching wiring connectors is illustrated in Figure ~\ref{fig:strategy} as an example of the strategies.

While using CAD models in simulations offers geometric reference, simulations lack real-time updates and interaction capabilities. The DT continuously synchronizes the robot and EVB positions, compensating for misalignments and enabling predictive collision avoidance. XR further enhances teleoperation by providing an intuitive, realistic interface for the disassembly programming while keeping operators safely outside the hazardous area. The combination of DT and XR improves safety, adaptability, and operational efficiency in EVB disassembly.

\subsection{Bidirectional communication}
\label{subsec:bidirectionalcommunication}
The data flow between the physical system and the DT is required to update the robot joint and TCP positions as well as the EVB position and orientation. The KUKA RSI commercial add-on is used on the robot controller to update the robot joint positions, and an open-source Kuka external control software development kit~\cite{kuka_industrial_experimental_2024} is used on the control PC.

\subsection{Human and Robot Capabilities in Hybrid EVB Disassembly}
\label{subsec:humanrobotcongnitive}
Achieving fully autonomous EVB disassembly is challenging due to design variability, lack of standardization, and permanent mechanical joints. While automation executes predefined sequences, human cognition is essential for adaptability and decision-making. The proposed approach integrates human-in-the-loop teleoperation to enhance flexibility and efficiency.

DT assists in planning disassembly, but teleoperators decide strategies for unknown battery types and decide tool usage. The XR interface enables real-time monitoring and intervention, ensuring precise execution without direct exposure to hazards. When unexpected issues arise, such as jammed screws or misaligned components, teleoperators can diagnose and reconsider operations as needed.

The presented hybrid approach improves safety, efficiency, and adaptability, addressing robotic limitations. Future research can further automate perception, decision-making, and recovery processes to reduce human intervention while maintaining the system’s flexibility.

\section{RESULTS AND DISCUSSION}
\label{sec:resultsdiscussion}

\begin{figure*}[!tb]
  \centering
  \includegraphics[width=2\columnwidth]{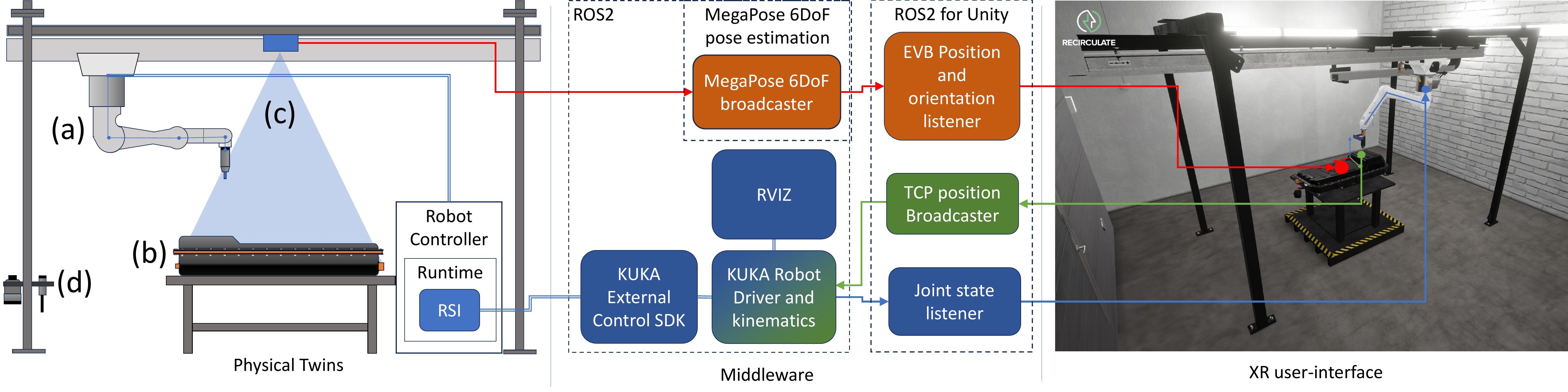}
  \caption{System architecture (a) Robot arm and linear track, (b) EVB, (c) RGB camera, (d) Stationary tool holder and resting tools.}
  \label{fig:architecture}
\end{figure*}

The outcome of the presented work is the concept of a teleoperation system designed to automate the disassembly and sorting of EVBs. The concept can be divided into DTs and XR user interface, presented in Sections~\ref{subsec:digitaltwins} and~\ref{subsec:xruserinterface}. While the disassembly of an EVB using an industrial robot arm was presented as a case example, the concept is scalable to various products and robot types; the only requirements are the CAD model of the product and the robot's ROS compatibility. 
By introducing an automated solution for disassembling and sorting EVB, the concept increases the profitability and safety of processing end-of-life products. The software architecture is detailed in Figure~\ref{fig:architecture}.

\subsection{Digital twins}
\label{subsec:digitaltwins}

The KUKA robot kinematic descriptions ROS-package~\cite{kroshukukadescription} provides the kinematic properties for trajectory calculations of the robot digital twin, matching the properties and limitations of the physical twin. The KUKA robot driver component receives the requested TCP position from the user interface and outputs the physical robot joint positions to the user interface. The Robot Visualization (RVIZ) ROS package generates collision-free trajectories to reach the requested positions. The external control SDK  provides real-time robot control using the RSI add-on. 

The position and orientation of the EVB are updated in real time using the MegaPose and an RGB camera. Based on the CAD model and video input, the MegaPose library identifies the location and orientation of the EVB in the disassembly and sorting cell. A ROS broadcaster component was written for the Megapose to broadcast the position and orientation of the EVB to Unity and ROS.  

\subsection{XR User interface}
\label{subsec:xruserinterface}
\begin{figure}[!tb]
  \centering
  \framebox{\parbox{3in}{\includegraphics[width=3in]{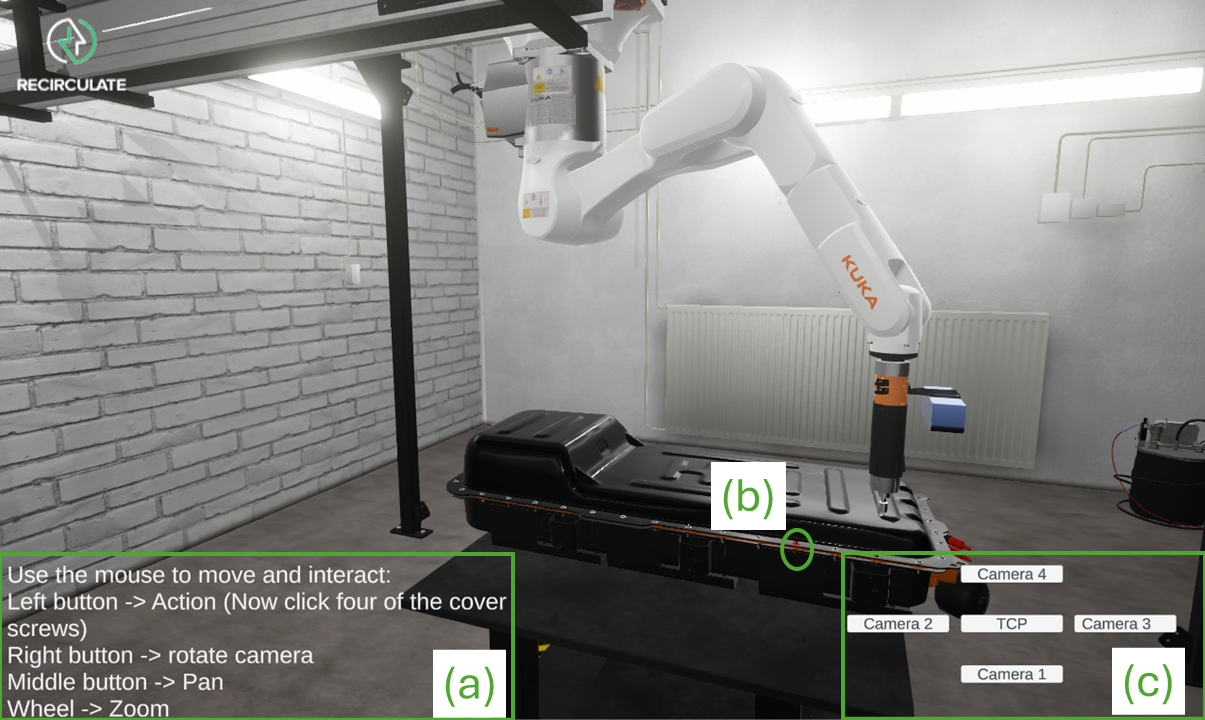}}}
  \caption{The XR user interface (a) Guidance, (b) screw activated by pointing and clicking, (c) camera preset buttons. }
  \label{fig:userinterface}
\end{figure}

The XR user interface enables the teleoperator to detach components of the EVB by pointing and clicking using the left mouse button to program the robot for the first time for an unknown EVB. The EVB disassembly teleoperation can be performed using a desktop or an immersive user interface. 
ROS2 for Unity add-on enables the ROS messaging protocol for the user interface. The TCP position broadcaster sends the robot TCP position requests to the KUKA robot driver on the ROS middleware, and the joint state listener updates the robot position in the user interface. EVB position and orientation listener updates the position and orientation of the EVB in the user interface. The user interface enables saving and rerunning the recorded TCP positions for the automated disassembly of previously programmed EVBs. The teleoperator can change the viewpoint to any location and orientation, zoom, and activate components for detaching by pointing and clicking. Figure~\ref{fig:userinterface} illustrates the user interface.

\subsection{Cost-benefit analysis}
\label{subsec:costbenefit}
A cost-benefit analysis justifies the industrial relevance of the presented approach. The presented EVB was manually disassembled, and the times for different disassembly phases were recorded to compare the efficiency to the robotized disassembly. Table~\ref{tab:comparision} provides details on the time required for the task during the pack-to-module level disassembly of the EVB. It is worth noting that due to safety regulations, manual disassembly requires two human workers.

\begin{table}[!tb]
    \centering
    \caption{Comparison of manual and robotized disassembly. }
    \label{tab:comparision}
    \begin{tabular}{|c||c||c|} \hline 
         Phase&  Manual time& Robotized time\\ \hline 
         Cover screw removal&  645 sec& 75 sec\\ \hline 
         Tool change&  ----& ----\\ \hline 
         Battery cover removal&  6 sec& 12 sec\\ \hline 
         Tool Change&  ----& ----\\ \hline 
         Wiring connectors detach&  71 sec& 240 sec\\ \hline 
         Tool Change&  ----& ----\\ \hline 
         Battery module screws removal&  76 sec& 92 sec\\ \hline 
         Tool change&  ----& ----\\ \hline 
         Total:&  258 sec& 474 sec\\ \hline
    \end{tabular}
\end{table}

The longer time for robotized wiring connector detachment highlights the dexterity of human hands; human hands can unlatch and pull the connector plug out of the socket with a single grip. On the contrary, the rigid construction of the robot gripper requires multiple linear movements to detach a single connector. Table~\ref{tab:costbreakdown} details the investments for the robotized approach.

\begin{table}[!tb]
    \centering
    \caption{Cost breakdown of the robotized EVB disassembly cell. }
    \label{tab:costbreakdown}
    \begin{tabular}{|c||c|} \hline 
         Item& Price\\ \hline 
         Robot and linear track& 39 500 €\\ \hline 
         Steel frame& 2600 €\\ \hline 
         Automatic tool changer& 2808 €\\ \hline 
         Vacuum gripper& 1140 €\\ \hline 
         Wiring connector detach gripper& 2300 €\\ \hline 
         EVB reverse design& 1100 €\\ \hline 
         Digital twin creation& 8300 €\\ \hline 
         User interface programming& 5600 €\\ \hline 
         Total:& 108 256 €\\ \hline
    \end{tabular}
\end{table}
The proposed concept's benefits include minimizing risks to human workers, which is enabled by remote programming and monitoring of the disassembly process. Manual disassembly requires two workers per task, doubling the labor cost compared to a robot.  

The challenges are the higher upfront investments compared to manual disassembly, the limited availability of detailed CAD models, and the slower speed of the robot for tasks requiring dexterity. Reverse engineering the variety of EVBs is time-consuming. Centria's professional engineers spent 32 working hours reverse engineering the PHEV battery presented in this publication.      

The investment becomes profitable after the first year of operation by reducing labor dependency.  Equation~\ref{eq:roi} presents an estimate of the Return of the Investment (ROI).

\begin{equation} \label{eq:roi}
\text{$ROI$} = \frac{\text{$NS$} - \text{$IC$}}{\text{$IC$}} = \frac{\text{100000 €} - \text{108256 €}}{\text{108256 €}} = -7.63 \% 
\end{equation}
\noindent where $ROI$ is the return of investment, $NS$ is net savings and $IC$ is the investment costs.  

While the initial ROI is negative due to upfront investment, labor dependency cost reductions and safety improvements are expected to offset this within approximately one year of operation. Furthermore, as more disassembly sequences are recorded, future automation will further enhance efficiency and scalability.

\subsection{Pilot study}
\label{subsec:piloting}

The proposed system's user interface was made available online for a pilot study to validate its functionality and evaluate its usability. A web link to the virtual user interface was shared on social media to attract pilot users. Thirty volunteers piloted the system and provided feedback; the pilot session consisted of unscrewing and removing the EVB cover, detaching wiring connectors and assemblies, and removing the battery modules and coolant piping.

After the pilot participants had completed the disassembly of the EVB, they were provided with a link to a voluntary online survey. User feedback was collected using a System Usability Scale (SUS) survey on a five-point Likert scale~\cite{brooke1996}. In addition to the standard SUS questions, one question was added to define the pilot users' previous knowledge in robotics and an optional free comment. A summary of the user feedback is presented in Table~\ref{tab:feedback}.

\begin{table}[htbp]
\caption{Results and analysis of the survey.}
\label{tab:feedback}
\centering
\renewcommand{\arraystretch}{1.5} 
\resizebox{\linewidth}{!}{%
\begin{tabular}{|l||p{0.9cm}||p{0.9cm}||p{0.8cm}||p{0.8cm}||p{0.9cm}|}
\hline
\textbf{Question}  
& \centering\textbf{Strongly\\ disagree}  
& \centering\textbf{Disagree} 
& \centering\textbf{Neutral}  
& \centering\textbf{Agree}  
& \textbf{\centering Strongly\\ agree}  \\ 
\hline
\begin{tabular}{@{}l@{}}1. I think that I would like to use this\\ system frequently.\end{tabular} 
& 3.1\%& 21.9\%& 31.3\%& \textbf{40.6\%}& 3.1\%\\ 
\hline
\begin{tabular}{@{}l@{}}2. I found the system unnecessarily complex.\end{tabular} 
& 18.7\%& \textbf{50.0\%}& 21.9\%& 6.3\%& 3.1\%\\  
\hline
\begin{tabular}{@{}l@{}}3. I thought the system was easy to use. \end{tabular} 
& 0.0\%& 20.0\%& 16.7\% & \textbf{43.3\%} & 20.0\%  \\  
\hline
\begin{tabular}{@{}l@{}}4. I think that I would need the support of\\ a technical person to be able to use this system. \end{tabular} 
& 25.8\%  & \textbf{58.1}\%& 6.4\%& 9.7\%& 0.0\%\\  
\hline
\begin{tabular}{@{}l@{}}5. I found the various functions in this system\\ were well integrated.\end{tabular} 
& 0.0\%& 20.0\%& 13.3\%& \textbf{66.7\%}& 0.0\%\\  
\hline
\begin{tabular}{@{}l@{}}6. I thought there was too much inconsistency\\ in this system.\end{tabular} 
& 16.7\% & \textbf{50.0\%} & 16.6\%& 10.0\%& 6.7\%\\  
\hline
\begin{tabular}{@{}l@{}}7. I would imagine that most people would\\ learn to use this system very quickly. \end{tabular} 
& 0.0\%  & 9.7\% & 9.7\% & \textbf{51.6\%} & 29.0\% \\  
\hline
\begin{tabular}{@{}l@{}}8. I found the system very cumbersome to use. \end{tabular} 
& 9.7\%  & \textbf{41.9\%} & 12.9\% & 25.8\% & 9.7\% \\  
\hline
\begin{tabular}{@{}l@{}}9. I felt very confident using the system. \end{tabular} 
& 3.4\%  & 23.3\% & \textbf{33.3\%} & \textbf{33.3\%} & 6.7\% \\  
\hline
\begin{tabular}{@{}l@{}}10. I needed to learn a lot of things before\\ I could get going with this system. \end{tabular} 
& 28.1\%  & \textbf{43.8\%} & 12.5\% & 15.6\% & 0.0\% \\  
\hline
\begin{tabular}{@{}l@{}}I feel that I am an expert in robotics. \end{tabular} 
& \textbf{50.0\%} & 16.0\% & 17.0\% & 0.0\% & 17.0\% \\  
\end{tabular}}
\end{table}
\vspace{-0.82cm}
\begin{table}[htbp]
\resizebox{\columnwidth}{!}{%
\begin{tabular}{|l|l|}
\hline
\multirow{3}{*}{Free comments} &
  \begin{tabular}[c]{@{}l@{}}1. It is a pity that it does not work with the laptop mouse touchpad. With\\ a proper mouse I finally managed to dismantle the battery successfully.\end{tabular} \\ \cline{2-2} 
 &
  \begin{tabular}[c]{@{}l@{}}2. Playing around with the mouse is cumbersome,\\  causing a lot of misclicks and wandering outside the working area.\end{tabular} \\ \cline{2-2} 
 &
  \begin{tabular}[c]{@{}l@{}}3. I tested the interface using my laptop touchpad, \\ which made some functions like zooming and rotating awkward.\end{tabular} \\ \hline
\end{tabular}%
}
\end{table}

The analysis of SUS scores shows a mean SUS score of 65.4, which is considered a good value. The first question \textit{I think that would like to use this system frequently} in this study is misleading. Asking \textit{I think I would like to use this system for EVB disassembly more frequently than manually disassembling EVBs} would have resulted in a more favorable result. The answers to questions 7 and 10 indicate that learning how to use and using the system is easy. 

While most participants think the system functionalities are well integrated and convenient to use, about 35\% feel the system is cumbersome. The mentions in the free comments likely reveal the root cause for this: using the mouse middle and right buttons for the pan and tilt functions is not natural to most of the users, as many of the users might not have had an external mouse connected to their device during the piloting. However, in the production environment, the teleoperator would have more convenient dedicated devices to control the process, such as a joystick, a 3D space mouse, and an AR/VR headset.         

The added question reveals that only half of the participants consider themselves as experts in robotics. This relates to the fact that only 40\% felt confident using the system, as they controlled an industrial robot and disassembled an EVB for the first time.

\section{CONCLUSIONS}
\label{sec:conclusions}
This paper presents a DT-based teleoperation approach for EVB disassembly, enabling safe and efficient remote-controlled disassembly and sorting. By integrating human cognitive capabilities with robotic automation, the system allows operators to create disassembly sequences while using the precision and repeatability of industrial robots. The safety of human workers is ensured by enabling teleoperators to control and monitor the disassembly process outside of the hazardous area. 

The feasibility of the presented approach relies on the accuracy of CAD models and of the pose estimation through MegaPose. Unfortunately, currently the KUKA ROS package~\cite{kroshukukadescription} does not support controlling additional axes, which prevented hardware-in-the-loop experiments to validate this accuracy. Therefore, software-in-the-loop support was implemented to test the concept at the software level only. The proposed system can be easily scaled up to enable the simultaneous disassembly of multiple and different batteries within a disassembly line.\\
Future work will further automate decision-making processes, improve real-time adaptability and user interface, controls, and conduct hardware-in-the-loop experiments to validate the proposed approach in an industrial setting.

\addtolength{\textheight}{-12cm}   





\section*{ACKNOWLEDGMENT}
This research was supported by the European Climate, Infrastructure and Environment Executive Agency (CINEA) under the European Union’s Horizon Europe Research and Innovation Programme grant number 101103972.


\bibliographystyle{ieeetr}
\bibliography{root}

\begin{thebibliography}{10}

\bibitem{fernandeznavarro2012}
P.~Fernández-Navarro, J.~García-Pérez, R.~Ramis, E.~Boldo, and G.~López-Abente, ``Proximity to mining industry and cancer mortality,'' {\em Science of The Total Environment}, vol.~435-436, pp.~66--73, 2012.

\bibitem{ramos2008}
W.~Ramos, C.~Galarza, G.~Ronceros, F.~De~Amat, M.~Teran, L.~Pichardo, D.~Juarez, R.~Anaya, A.~Mayhua, J.~Hurtado, and A.~Ortega-Loayza, ``{Noninfectious dermatological diseases associated with chronic exposure to mine tailings in a Peruvian district},'' {\em British Journal of Dermatology}, vol.~159, pp.~169--174, 07 2008.

\bibitem{Priyono2016}
A.~Priyono, W.~Ijomah, and U.~Bititci, ``Disassembly for remanufacturing: A systematic literature review, new model development and future research needs,'' {\em Journal of Industrial Engineering and Management}, vol.~9, no.~4, pp.~899--932, 2016.

\bibitem{Kaarlela2024}
T.~Kaarlela, E.~Villagrossi, A.~Rastegarpanah, A.~San-Miguel-Tello, and T.~Pitkäaho, ``Robotised disassembly of electric vehicle batteries: A systematic literature review,'' {\em Journal of Manufacturing Systems}, vol.~74, pp.~901--921, 2024.

\bibitem{Zang2022}
Y.~Zang and Y.~Wang, ``Robotic disassembly of electric vehicle batteries: {An} overview,'' in {\em 2022 27th International Conference on Automation and Computing (ICAC)}, pp.~1--6, 2022.

\bibitem{Villagrossi2023}
V.~Enrico and D.~Tito, ``Robotics for electric vehicles battery packs disassembly towards sustainable remanufacturing,'' {\em Journal of Remanufacturing}, vol.~13, no.~3, pp.~355--379, 2023.

\bibitem{Goertz1952}
R.~C. Goertz, ``Fundamentals of general-purpose remote manipulators,'' in {\em Nucleonics}, 1952.

\bibitem{Szczurek2023}
K.~A. Szczurek, R.~M. Prades, E.~Matheson, J.~Rodriguez-Nogueira, and M.~D. Castro, ``Multimodal multi-user mixed reality human–robot interface for remote operations in hazardous environments,'' {\em IEEE Access}, vol.~11, pp.~17305--17333, 2023.

\bibitem{Ito2019}
M.~Ito, Y.~Funahara, S.~Saiki, Y.~Yamazaki, and Y.~Kurita, ``Development of a cross-platform cockpit for simulated and tele-operated excavators,'' {\em Journal of Robotics and Mechatronics}, vol.~31, no.~2, pp.~231--239, 2019.

\bibitem{poysari2023}
S.~P\"oys\"ari, T.~Kaarlela, M.~Dianatfar, and M.~Lanz, ``Educational {Teleoperation} {Platform} for {Heavy} {Industrial} {Robotics} as a {Learning} {Environment},'' in {\em Proceedings of the 13th Conference on Learning Factories}, (Rochester, NY), June 2023.

\bibitem{Hathaway2023}
J.~Hathaway, A.~Shaarawy, C.~Akdeniz, A.~Aflakian, R.~Stolkin, and A.~Rastegarpanah, ``Towards reuse and recycling of {Lithium-Ion} batteries: {Tele-robotics} for disassembly of electric vehicle batteries,'' {\em Frontiers in Robotics and AI}, vol.~10, 2023.

\bibitem{Garg2021}
G.~Garg, V.~Kuts, and G.~Anbarjafari, ``Digital twin for fanuc robots: Industrial robot programming and simulation using virtual reality,'' {\em Sustainability}, vol.~13, no.~18, 2021.

\bibitem{Xin2021}
X.~Li, B.~He, Y.~Zhou, and G.~Li, ``Multisource model-driven digital twin system of robotic assembly,'' {\em IEEE Systems Journal}, vol.~15, no.~1, pp.~114--123, 2021.

\bibitem{labbe2022}
Y.~Labb\'e, L.~Manuelli, A.~Mousavian, S.~Tyree, S.~Birchfield, J.~Tremblay, J.~Carpentier, M.~Aubry, D.~Fox, and J.~Sivic, ``Megapose: 6d pose estimation of novel objects via render compare,'' in {\em Proceedings of the 6th Conference on Robot Learning (CoRL)}, 2022.

\bibitem{kroshukukadescription}
A.~Svastits, ``kuka\_robot\_descriptions.'' https://github.com/kroshu/kuka\_robot\_descriptions, 2024.
\newblock Accessed 14 June 2024.

\bibitem{kuka_industrial_experimental_2024}
{ROS developers}, ``ros-industrial/kuka\_experimental.'' https://github.com/ros\-industrial/kuka\_experimental, 2024.
\newblock Accessed 14 June 2024.

\bibitem{brooke1996}
J.~Brooke {\em et~al.}, ``{SUS-A} quick and dirty usability scale,'' {\em Usability evaluation in industry}, vol.~189, no.~194, pp.~4--7, 1996.

\end{thebibliography}

\end{document}